\documentclass{article} 
\usepackage{iclr2025_conference,times}

\usepackage{amsmath,amsfonts,bm}

\def\eqref#1{equation~\ref{#1}}

\def\1{\bm{1}}

\DeclareMathAlphabet{\mathsfit}{\encodingdefault}{\sfdefault}{m}{sl}
\SetMathAlphabet{\mathsfit}{bold}{\encodingdefault}{\sfdefault}{bx}{n}

\usepackage{tikz}
\usepackage{hyperref}
\usepackage{url}
\usepackage{graphicx}
\usepackage{amsmath}
\usepackage{amssymb}
\usepackage{booktabs}
\usetikzlibrary{arrows.meta, positioning, shapes.geometric, calc, backgrounds, shadows, fit}

\definecolor{elsevierblue}{RGB}{0, 115, 152}
\definecolor{neuralorange}{RGB}{230, 126, 34}
\definecolor{guardgreen}{RGB}{39, 174, 96}
\definecolor{logicgray}{RGB}{245, 245, 245}
\definecolor{coreblue}{HTML}{0072B2}

\title{Modal Logical Neural Networks for\\ Financial AI}

\author{Antonin Sulc \\
Lawrence Berkeley National Lab\\
Berkeley, CA, USA \\
\texttt{asulc@lbl.gov} \\
}

\iclrfinalcopy 
\begin{document}

\maketitle

\begin{abstract}
The financial industry faces a critical dichotomy in AI adoption: deep learning often delivers strong empirical performance, while symbolic logic offers interpretability and rule adherence expected in regulated settings. We use Modal Logical Neural Networks (MLNNs) as a bridge between these worlds, integrating Kripke semantics into neural architectures to enable differentiable reasoning about necessity, possibility, time, and knowledge. We illustrate MLNNs as a differentiable ``Logic Layer'' for finance by mapping core components, Necessity Neurons ($\Box$) and Learnable Accessibility ($A_\theta$), to regulatory guardrails, market stress testing, and collusion detection. Four case studies show how MLNN-style constraints can promote compliance in trading agents, help recover latent trust networks for market surveillance, encourage robustness under stress scenarios, and distinguish statistical belief from verified knowledge to help mitigate robo-advisory hallucinations.
\end{abstract}

\section{Introduction: The Logic-Learning Gap in Finance}

The deployment of Artificial Intelligence in finance has reached a critical inflection point. While Deep Reinforcement Learning (RL) and Large Language Models (LLMs) have demonstrated exceptional skill in identifying statistical arbitrage and generating market commentary, they often lack explicit representations of declarative constraints. These models typically operate as black-box probability mappers, optimizing for reward signals without a native understanding of the rigid, declarative constraints that govern the financial system. This methodological gap creates significant deployment risks. An RL trading bot may discover that "washing" trades—selling and immediately repurchasing to manipulate volume or tax liability—is a highly profitable strategy, having no internal representation of the regulatory axiom that forbids it. Conversely, traditional Symbolic AI systems can enforce these rules but can be brittle when adapting to the stochastic, high-dimensional nature of modern market data.

To bridge this divide, we propose the application of Modal Logical Neural Networks (MLNNs)~\cite{sulc2025modal}  to financial domains. Unlike standard logical guardrails that check if a static fact is true, MLNNs\footnote{Code available at \url{https://github.com/sulcantonin/torchmodal}; \texttt{pip install torchmodal}} integrate Kripke semantics directly into the neural architecture. This allows the model to reason not just about the current state of the market, but about the structure of "possible worlds"—a useful capability for robust Financial AI.

This paper presents four finance-facing scenarios where modal reasoning can support safety-critical capabilities that are often awkward to express in purely statistical objectives: temporal compliance, collusion detection, tail-risk -- robust optimization, and interpretable contract review, positioning modal logic as a practical safety layer for AI in banking and investment.

\begin{figure*}[t]
\centering
\resizebox{1.0\linewidth}{!}{
\begin{tikzpicture}[
    font=\sffamily,
    >=Stealth,
    section_title/.style={font=\Large\bfseries, color=elsevierblue, anchor=north west},
    box/.style={draw, thick, rounded corners=3pt, align=center, fill=white, drop shadow={opacity=0.15}},
    soft_node/.style={rectangle, rounded corners=0.8em, fill=white, align=center, inner sep=5pt, font=\small},
    core_node/.style={rectangle, rounded corners=0.8em, fill=coreblue!10, align=center, inner sep=5pt, font=\small},
    loss_soft/.style={rectangle, rounded corners=0.8em, fill=purple!10, align=center, inner sep=5pt, font=\small},
    swoosh_flow/.style={->, ultra thick, gray!30, shorten >=2pt, shorten <=2pt},
    swoosh_gradient/.style={->, line width=2pt, neuralorange, shorten >=3pt, shorten <=3pt},
    scenario_box/.style={box, minimum width=4.8cm, minimum height=1.4cm, font=\small}
]

\node[section_title] at (0, 3.2) {I. Kripke Semantics};
\node[box, fill=logicgray!40, minimum width=5.5cm, minimum height=4.2cm] (KripkeFrame) at (2.75, 0.5) {};

\node[draw, circle, ultra thick, minimum size=1cm, fill=white] (w1) at (1.2, 0.8) {$t$};
\node[draw, circle, ultra thick, minimum size=1cm, fill=white] (w2) at (3.8, 1.5) {$t{+}1$};
\node[draw, circle, ultra thick, minimum size=1cm, fill=red!15] (w3) at (3.8, -0.3) {$w_c$};

\draw[->, ultra thick, elsevierblue] (w1) to [bend left=20] node[midway, above, sloped, font=\small] {Temporal} (w2);
\draw[->, ultra thick, gray, dashed] (w1) to [bend right=20] node[midway, below, sloped, font=\small] {Stress} (w3);

\node[below=0.3cm of KripkeFrame, align=center, font=\small\bfseries] {
    Modal Operators\\[2pt]
    {\Large $\Box$} (Necessity) \\
    {\Large $\Diamond$} (Possibility)
};

\node[section_title] at (6.5, 3.2) {II. Learning Modes};

\node[font=\small\bfseries, color=elsevierblue, anchor=west] at (6.5, 2.4) {(A) Deductive: Fixed $R$};
\node[soft_node, anchor=west] (PropNN) at (6.8, 1.6) {\textbf{Neural Net}\\Updates $V$};
\node[core_node, right=0.4cm of PropNN] (CoreA) {\textbf{Modal Logic}\\States $V$};
\node[loss_soft, right=0.4cm of CoreA] (LossA) {\textbf{Loss}\\$L_{\text{contra}}$};

\draw[swoosh_flow] (PropNN) -- (CoreA);
\draw[swoosh_flow] (CoreA) -- (LossA);
\draw[swoosh_gradient] (LossA.south) to[out=250, in=290, looseness=0.5] (PropNN.south);

\node[font=\small, color=gray, below=0.05cm of CoreA] {e.g., Temporal flow};

\node[font=\small\bfseries, color=elsevierblue, anchor=west] at (6.5, -0.2) {(B) Inductive: Learn $A_\theta$};
\node[soft_node, anchor=west] (RelNN) at (6.8, -1.5) {\textbf{Relation Net}\\Learns $A_\theta$};
\node[core_node, right=0.4cm of RelNN] (CoreB) {\textbf{Modal Logic}\\Relations};
\node[loss_soft, right=0.4cm of CoreB] (LossB) {\textbf{Loss}\\$L_{contra}$};

\draw[swoosh_flow] (RelNN) -- (CoreB);
\draw[swoosh_flow] (CoreB) -- (LossB);
\draw[swoosh_gradient] (LossB.south) to[out=250, in=290, looseness=0.5] (RelNN.south);

\node[font=\small, color=gray, below=0.05cm of CoreB] {e.g., Trust networks};

\node[section_title] at (14.5, 3.2) {
};

\node[scenario_box, draw=elsevierblue, fill=elsevierblue!5] (S1) at (17.2, 2.0) {
    \textbf{1. Wash Sale Guardrail}\\[2pt]
    \textit{Deductive compliance}
};

\node[scenario_box, draw=neuralorange, fill=neuralorange!5] (S2) at (17.2, 0.2) {
    \textbf{2. Collusion Detection}\\[2pt]
    \textit{Inductive surveillance}
};

\node[scenario_box, draw=guardgreen, fill=guardgreen!5] (S3) at (17.2, -1.6) {
    \textbf{3. Crash-Proof Portfolio}\\[2pt]
    \textit{Robust optimization}
};

\node[scenario_box, draw=purple!70, fill=purple!5] (S4) at (17.2, -3.4) {
    \textbf{4. Safe Signer (CUAD)}\\[2pt]
    \textit{Interpretable risk}
};


\node[box, fill=elsevierblue!10, draw=elsevierblue, minimum width=19cm, minimum height=0.8cm] at (10, -4.6) {
    \textbf{Optimization:} $L_{total} = L_{\text{task}} + \beta \cdot L_{\text{contra}}$ \quad | \quad $M = \langle W, R, V \rangle$ differentiable
};

\draw[gray!20, line width=1.2pt] (5.8, -4.0) -- (5.8, 2.8);
\draw[gray!20, line width=1.2pt] (14.0, -4.0) -- (14.0, 2.8);
\end{tikzpicture}}
\caption{\textbf{Modal Logical Neural Networks for Finance.} \textbf{Left:} Kripke semantics with possible worlds (temporal: $t, t+1$; stress: crash world $w_c$). \textbf{Middle:} Two learning modes—(A) Deductive: fixed accessibility $R$ (e.g., time flow), (B) Inductive: learned $A_\theta$ (e.g., trust networks). Modal operators $\Box$ (necessity: all worlds) and $\Diamond$ (possibility: $\geq$1 world) enable differentiable logical reasoning. \textbf{Right:} Four financial applications leveraging these modes for compliance, surveillance, risk management, and interpretability.}
\label{fig:teaser}
\end{figure*}
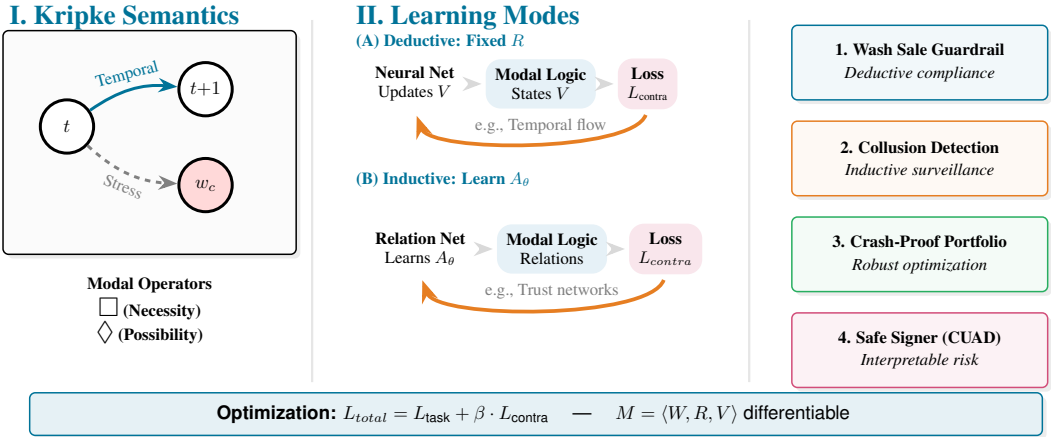

\subsection{Related Work}

Neuro-symbolic architectures like Logic Tensor Networks \citep{serafini2016logic, badreddine2022logic} and Logical Neural Networks \citep{riegel2020logical} successfully integrate deep learning with first-order reasoning, while DeepProbLog \citep{manhaeve2018deepproblog} incorporates probabilistic logic. However, these frameworks do not explicitly model \textit{Kripke semantics}, which are useful for capturing temporal dynamics and epistemic states in financial settings. Unlike formal verification methods that operate post-hoc \citep{katz2017reluplex}, MLNNs enforce constraints \textit{during} training via differentiable loss terms, extending the ``semantic loss'' \citep{xu2018semantic} and soft-logic \citep{fischer2019dl2} paradigms to modal operators. This can yield models with more explicit, auditable structure \citep{rudin2019stop}, providing transparency without sacrificing neural expressiveness.

\section{The Differentiable Logic Layer in Finance}

An MLNN fundamentally transforms a neural network into a differentiable Kripke model defined by the tuple $M = \langle W, R, V \rangle$, see Figure~\ref{fig:teaser}. To understand its utility in finance, we must first deconstruct these components and how they enable a transition from static statistical mapping to dynamic logical reasoning.

Possible Worlds $W$ represent distinct states: future timesteps (temporal reasoning), economic scenarios (stress testing), or agent perspectives (epistemic reasoning). The Valuation function $V$ assigns continuous truth values to propositions within each world.

The Accessibility Relation $R$ (or learnable $A_\theta$) defines which worlds are relevant. In {Deductive mode}, $R$ is fixed (e.g., temporal flow for wash sale rules). In {Inductive mode}, $A_\theta$ is learned from data to discover hidden structure (e.g., trust networks for collusion detection)

Modal Neurons aggregate over worlds: Necessity ($\Box$) requires truth in \textit{all} accessible worlds (portfolio solvency in every stress scenario); Possibility ($\Diamond$) requires truth in \textit{at least one} world (existence of valid exit paths).

Doxastic/Epistemic Logic distinguishes Belief $B$ (statistical confidence) from Knowledge $K$ (verified in accessible worlds), preventing robo-advisory hallucinations. Training minimizes Logical Contradiction Loss $L_{\text{contra}}$, penalizing axiom violations.

\section{Scenarios}

We now explore four scenarios demonstrating how to map financial problems into this modal framework. Scenarios 1--3 present proof-of-concept simulations illustrating the MLNN's logical mechanics; Scenario 4 provides experimental validation on real data with baseline comparisons.

\subsection{Scenario 1: Agentic Compliance - The "Wash Sale" Guardrail}

\textbf{Concept:} Fixed Temporal Logic ($\Box$) constrains RL Agents.

In the domain of autonomous trading, Reinforcement Learning (RL) agents frequently discover that "washing" trades—selling a losing position to harvest a tax deduction and immediately buying it back—is a profitable strategy. While financially optimal in the short term, this violates time-bound regulations. To address this, we wrap the RL agent in a Temporal MLNN, where "possible worlds" represent future time steps $W = \{t, t+1, \dots, t+30\}$. We enforce a modal axiom stating that a `SellAtLoss` action implies that a `Buy` action cannot occur in any accessible future world 

$$\Box_{[t, t+30]} \neg \text{Buy}$$. 

When the agent proposes a sell order, the Necessity Neuron ($\Box$) scans the accessible future. If the agent's policy simultaneously assigns high probability to a buy order within the restricted window, the neuron detects a logical contradiction. The gradient flows backward from the future world to the current state, treating the regulation as a differentiable constraint during training.

In simulation, a standard RL agent exploited wash sales (\texttt{[BBSBBBBBBB]}: Profit 17.96, 1 Violation). The MLNN agent ($\beta$ annealed 0.0→2.0) discovered compliant strategy \texttt{[BB.BBBBBBB]} (Profit 12.86, 0 Violations), demonstrating strategic reshaping rather than simple blocking.

\subsection{Scenario 2: Market Surveillance - The "Collusion Hunter"}

{The Concept:} {Using Inductive Logic to discover hidden cartels ($A_\theta$).}

Detecting market manipulation often requires identifying hidden coordination between theoretically independent actors. For example, in a "spoofing" ring, Trader A may place fake orders to drive prices up while Trader B profits by selling. To an observer looking at agents individually, this appears legal; the crime lies in the relationship. MLNNs can detect this by treating market participants as nodes in a graph where the accessibility relation $A_\theta$ represents "Latent Trust."

We hypothesize a collusion axiom: $\text{Spoof}(\text{Trader}_i) \implies \Diamond_{A_\theta} \text{Profit}(\text{Trader}_j)$. By operating in an inductive mode, the model updates the learnable accessibility matrix $A_\theta$ to minimize logical contradictions in the observed data. 

To validate this, we simulated a market with 5 traders where Trader 0 acted as a "Spoofer" and Trader 1 as a "Beneficiary," amidst random noise. We trained the MLNN to discover the network structure $A_\theta$ purely from the axiom. The model recovered the cartel structure, assigning a trust weight of 0.9997 to the link $T_0 \to T_1$ while suppressing all non-colluding links to 0.00 (resulting in a sparse "Social X-Ray" of the market). Crucially, the model learned to ignore random spoofing events by other traders, suggesting that the sparsity penalty ($L_{\text{sparsity}}$) can help filter noise and highlight more stable structural coordination.

\subsection{Scenario 3: Risk Management - "Crash-Proof" Portfolios}
In this example we show use of $\Box$ operator for robust Optimization.

Standard portfolio optimization maximizes expected returns based on probability-weighted averages. This approach can be risky for critical solvency constraints because it allows "tail risks" to be masked by high-probability normal outcomes. A portfolio might generate excellent returns on average but face total insolvency in a rare "Black Swan" event.

MLNNs address this by replacing statistical expectation with logical necessity. We explicitly define a set of stress worlds $W$ (e.g., $w_{\text{Normal}}$, $w_{\text{Crash}}$) and apply the Necessity Operator ($\Box$) to the solvency proposition: 

$$\text{Portfolio} \implies \Box_{W} (\text{Solvent}).$$ 

Unlike a standard loss function which sums errors, the $\Box$ neuron functions as a differentiable "min-pooling" operator, asserting that the portfolio is valid only if the solvency constraint holds in {every} accessible world, regardless of its probability.

Simulating Bond (2\% return) vs. Risky Asset (10\% return, -50\% in crash): classical optimizer allocated 100\% Risky (E[R]=6.99\%, crash value 0.50). MLNN with $\Box(\text{Value} \ge 0.90)$ allocated 95\% Bonds (E[R]=2.27\%, crash value 0.99), prioritizing survival over expected return.

\subsection{Scenario 4: The Safe Signer - Interpretable Contract Review}

Standard neural classifiers for legal documents suffer from \textit{opacity}: when a model outputs ``P(Safe) = 0.51,'' professionals receive no insight into \textit{why} the system is uncertain. We show that MLNNs can provide \textbf{decomposed, interpretable reasoning} that is often difficult to obtain from standard black-box classifiers.

\paragraph{Architecture and Setup.}
We implemented a ``Safe Signer'' agent on the Contract Understanding Atticus Dataset (CUAD)~\cite{hendrycks2103cuad}, using a multi-world Kripke structure with four risk levels: $w_0$ (Safe) through $w_3$ (Severe). A {Proposer Head} computes Belief $B$ from contract titles; an {Auditor Head} computes accessibility $A_\theta(w_i)$ to each risk world from clause text. Knowledge emerges via: $K = \Box(\text{Safe}) = \min_i[1 - A_\theta(w_i) \cdot \text{severity}_i]$. Full architectural details are provided in Appendix~\ref{app:safe_signer}.

\paragraph{Results.}
Table~\ref{tab:cuad_results} shows that while F1 scores are comparable (0.883), the MLNN achieved {100\% trap detection on our CUAD test split} on documents where standard titles mask hidden risks (vs. 96.6\% baseline), with a Belief-Knowledge gap of 0.995.

\begin{table}[h]
\centering
\begin{minipage}[t]{0.48\textwidth}
\centering
\caption{Performance on CUAD Test Set}
\label{tab:cuad_results}
\small
\begin{tabular}{@{}lccc@{}}
\toprule
\textbf{Model} & \textbf{F1} & \textbf{Trap} & \textbf{B-K} \\
\midrule
Baseline & 0.883 & 96.6\% & ,  \\
+ Threshold & 0.870 & 96.6\% & ,  \\
\textbf{MLNN} & \textbf{0.883} & \textbf{100\%} & \textbf{0.995} \\
\bottomrule
\end{tabular}
\end{minipage}%
\hfill
\begin{minipage}[t]{0.48\textwidth}
\centering
\caption{Interpretability on Ambiguous Cases}
\label{tab:cases}
\small
\resizebox{1.0\linewidth}{!}{
\begin{tabular}{@{}lccccl@{}}
\toprule
\textbf{Contract} & \textbf{Base.} & \textbf{$B$} & \textbf{$w_3$} & \textbf{$K$} & \textbf{Expl.} \\
\midrule
Joint Venture & 0.487 & 1.00 & 0.00 & 0.98 & Ver. Safe \\
IP Agreement & 0.508 & 1.00 & 0.88 & 0.12 & Sev. Risk \\
\bottomrule
\end{tabular}}
\end{minipage}
\end{table}
\paragraph{Interpretability Advantage.}
The MLNN's value emerges on ambiguous cases. On 29 samples (4.5\%) where the baseline produced confidence in the 0.25--0.75 range, the MLNN achieved 62.1\% accuracy versus 55.2\%, and crucially, \textit{explained why}. Consider two contracts with nearly identical baseline predictions ($P \approx 0.50$):

Both baselines say ``maybe''—the MLNN explains \textit{which component} triggered concern. Across the test set, the MLNN clustered into interpretable categories: 475 Verified Safe (high $K$), 155 Trap Detected (high $B$, low $K$), and 19 Uncertain (medium $K$). The learned temperature $\tau=0.02$  suggests strict modal enforcement is beneficial in this setup, aligning with the legal principle that one toxic clause can invalidate an otherwise standard agreement.

\section{Conclusion}

We introduced Modal Logical Neural Networks as a differentiable ``Logic Layer'' for financial AI, bridging statistical optimization and regulatory constraints. Through four scenarios, we demonstrated versatility across \textit{Deductive} mode (fixed temporal logic for compliance, stress-world solvency) and \textit{Inductive} mode (learned accessibility for collusion detection, epistemic uncertainty).

The Safe Signer experiment on CUAD provides an empirical illustration: 100\% trap detection on our test split (vs. 96.6\% baseline) with interpretable Belief-Knowledge decomposition. On ambiguous cases, the MLNN achieved 62.1\% accuracy versus 55.2\% while explaining uncertainty sources—essential for regulatory compliance where ``the neural network said 0.48'' is not auditable.

These results suggest modal logic is a practical tool for building financial AI systems that are safe, compliant, and auditable. Future work will extend to temporal reasoning over transaction sequences and multi-agent epistemic models for market surveillance.

\bibliography{iclr2025_conference}
\bibliographystyle{iclr2025_conference}

\appendix

\section{Safe Signer: Architecture Details}
\label{app:safe_signer}

\paragraph{Multi-World Kripke Structure.}
The Safe Signer extends the MLNN to four risk worlds with increasing severity:
\begin{itemize}
    \item $w_0$: Safe world (severity = 0.0)
    \item $w_1$: Minor risk world (severity = 0.3)
    \item $w_2$: Moderate risk world (severity = 0.6)
    \item $w_3$: Severe risk world (severity = 1.0)
\end{itemize}

\paragraph{Dual-Head Architecture.}
The model consists of two neural heads:
\begin{itemize}
    \item \textbf{Proposer Head}: Processes contract titles via self-attention and outputs Belief $B \in [0,1]$, representing the statistical prior that standard-looking documents are safe.
    \item \textbf{Auditor Head}: Processes clause text and outputs accessibility $A_\theta(w_i) \in [0,1]$ to each risk world.
\end{itemize}

\paragraph{Knowledge Computation.}
Knowledge is computed via the differentiable Necessity operator:
$$K = \Box(\text{Safe}) = \text{softmin}_\tau \left( \{1 - A_\theta(w_i) \cdot \text{severity}_i\}_{i=0}^{3} \right)$$
where $\text{softmin}_\tau(x) = -\tau \log \sum_i \exp(-x_i/\tau)$. The final Knowledge is capped by Belief: $K_{\text{final}} = \min(K, B)$.

\paragraph{Training Objective.}
The model is trained with a combined loss:
$$L = L_{\text{belief}} + L_{\text{risk}} + \lambda_1 L_{\text{contrastive}} + \lambda_2 L_{\text{axiom}}$$
where $L_{\text{contrastive}}$ enforces Belief-Knowledge separation on trap documents, and $L_{\text{axiom}}$ penalizes violations of modal axioms (e.g., $K \leq B$).

\paragraph{Explanation Categories.}
The MLNN's decomposition naturally clusters into interpretable categories:

\begin{table}[h]
\centering
\caption{MLNN Explanation Categories on CUAD Test Set}
\begin{tabular}{@{}lcp{5.5cm}@{}}
\toprule
\textbf{Category} & \textbf{Count} & \textbf{Explanation} \\
\midrule
Verified Safe & 475 & ``Safe across all risk worlds.'' \\
Trap Detected & 155 & ``Title standard, but clause opens risk world.'' \\
Uncertain & 19 & ``Moderate risk. Human review recommended.'' \\
\bottomrule
\end{tabular}
\end{table}

\paragraph{Hyperparameters.}
Embedding dimension: 128, Hidden dimension: 64, Attention heads: 4, Learning rate: 0.001, Epochs: 50, Batch size: 32, $\lambda_1 = 0.3$, $\lambda_2 = 0.2$.

\end{document}